\documentclass[10pt,twocolumn,letterpaper]{article}

\usepackage{cvpr}
\usepackage{times}
\usepackage{epsfig}
\usepackage{graphicx}
\usepackage{amsmath}
\usepackage{amssymb}
\usepackage{xcolor}
\usepackage{textcomp, gensymb}
\usepackage{subcaption}
\usepackage{url}
\usepackage{booktabs}
\usepackage{amsfonts} 
\usepackage{nicefrac}
\usepackage{times}
\usepackage{epsfig}
\usepackage{graphicx}
\usepackage{multirow}
\usepackage{bm}
\usepackage{bbm}
\usepackage{pifont}
\usepackage{xspace} 
\usepackage{graphicx,calc}
\usepackage{color}  
\usepackage{ifthen}
\usepackage{paralist}
\usepackage{times}
\usepackage{longtable}
\usepackage{colortbl}
\usepackage{nomencl}
\usepackage{siunitx}
\usepackage{csvsimple}
\usepackage{bm}
\usepackage{authblk}
\usepackage{adjustbox}
\usepackage{paralist}
\usepackage{pgfplots}
\usepackage{pgfplotstable}
\usepackage{float}

\pgfplotsset{compat=1.18}
\usetikzlibrary{patterns}

\usepackage[pagebackref=true,breaklinks=true,colorlinks,bookmarks=false]{hyperref}

\def\cvprPaperID{****} % *** Enter the CVPR Paper ID here

\newcommand{\todo}[1]{}
\renewcommand{\todo}[1]{{\color{red} TODO: {#1}}}

\ifcvprfinal\fi

\title{WayveScenes101: \\A Dataset and Benchmark for Novel View Synthesis in Autonomous Driving}

\author[*]{Jannik Zürn}
\author[*]{Paul Gladkov}
\author[ ]{Sofía Dudas}
\author[ ]{Fergal Cotter}
\author[ ]{Sofi Toteva}
\author[ ]{Jamie Shotton}
\author[ ]{Vasiliki Simaiaki}
\author[ ]{Nikhil Mohan}

\affil[ ]{Wayve}
\affil[*]{Equal contributions}

\begin{document}
\maketitle

%%%%%%%%% ABSTRACT
\begin{abstract}
We present WayveScenes101, a dataset designed to help the community advance the state of the art in novel view synthesis that focuses on challenging driving scenes containing many dynamic and deformable elements with changing geometry and texture. The dataset comprises 101 driving scenes across a wide range of environmental conditions and driving scenarios. The dataset is designed for benchmarking reconstructions on in-the-wild driving scenes, with many inherent challenges for scene reconstruction methods including image glare, rapid exposure changes, and highly dynamic scenes with significant occlusion. Along with the raw images, we include COLMAP-derived camera poses in standard data formats. We propose an evaluation protocol for evaluating models on held-out camera views that are off-axis from the training views, specifically testing the generalisation capabilities of methods. Finally, we provide detailed metadata for all scenes, including weather, time of day, and traffic conditions, to allow for a detailed model performance breakdown across scene characteristics. Dataset and code are available at \url{https://github.com/wayveai/wayve_scenes}.
\end{abstract}

%%%%%%%%% BODY TEXT
\section{Introduction}
\label{sec:intro}

Novel view synthesis methods have achieved remarkable success for scene reconstruction tasks across many applications~\cite{barron2021mip, muller2022instant, kerbl20233d}. Within the domain of autonomous driving, this technology potentially opens a new avenue for using synthetic data from reconstructed scenes both for evaluating driving models in simulation and for training them. Novel view synthesis models can be used to render realistic images from novel viewpoints which makes them a feasible alternative to other rendering methods.

\begin{figure}[H]
  \centering
  \includegraphics[width=\linewidth]{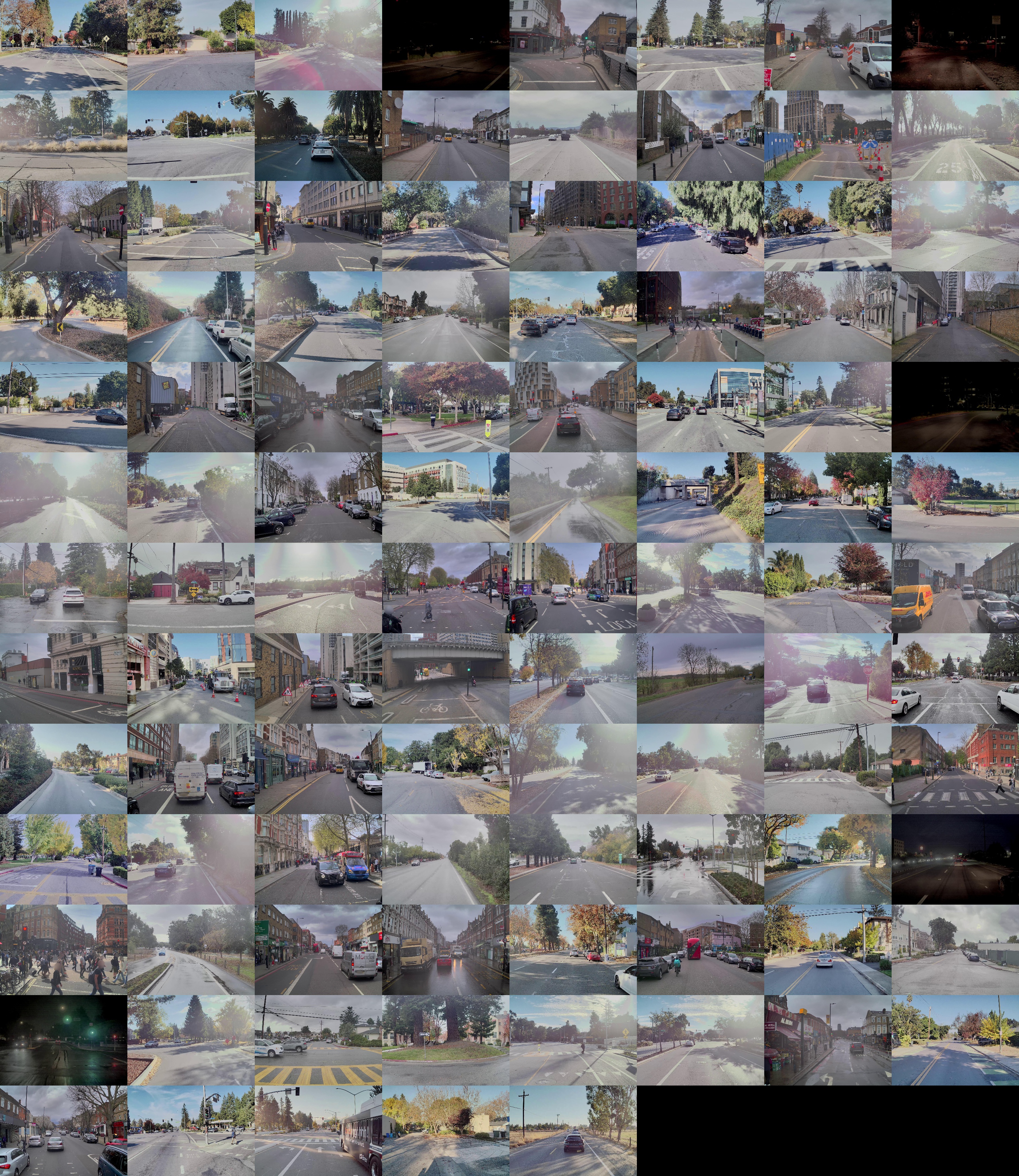}
  \caption{Overview of the 101 scenes in our dataset.}
  \label{fig:101_scenes}
\end{figure}

Prior datasets explicitly targeted on novel view synthesis focus on synthetic scenes or object-centric scenes with limited applicability to the challenges of driving~\cite{barron2021mip, hedman2018deep}. Such large-scale driving scenes present a number of challenges for novel view synthesis models that are not well captured in existing datasets. Firstly, they entail a variety of dynamic and possibly deformable objects, including pedestrians, cyclists, vehicles, trucks, and vegetation. Accurately reconstructing these scene parts is challenging for current methods. Furthermore, driving scenes can exhibit dynamic lighting conditions such as lens flare, reflections, and fast camera exposure changes, which are difficult to model. Finally, driving environments are usually recorded with camera rigs mounted on vehicles, resulting in a limited set of camera viewpoints which have a similar distance from the ground. This is particularly significant in scenes with many static or dynamic obstacles such as vehicles and pedestrians, which leads to a significant amount of occlusions, introducing additional challenges for reconstruction. Existing autonomous driving datasets~\cite{geiger2012we,Argoverse2, Sun_2020_CVPR} may contain large quantities of driving scenes but do not focus explicitly on scene diversity and their respective camera rig setups are not specifically targeted to evaluating novel view synthesis methods.

To address the aforementioned shortcomings of prior datasets, we present WayveScenes101, a large-scale dataset for novel view synthesis methods on driving scenes. Our dataset consists of 101 outdoor driving scenes with a length of 20 seconds each, featuring a wide range of environmental conditions and driving situations. Overall, our dataset contains 101,000 camera images and associated camera poses obtained from COLMAP~\cite{schoenberger2016sfm}.

To summarise, the dataset key features are:

\begin{compactitem}
    \item 101 diverse driving scenarios of 20 seconds each
    \item 101,000 images (101 scenes × 5 cameras × 20 seconds × 10 frames per second)
    \item Scene recording locations: US and UK
    \item Held-out evaluation camera for off-axis reconstruction quality measurement
    \item Scene metadata for targeted model evaluation
\end{compactitem}

\section{Related Work}
\label{sec:related}

We review existing datasets for scene reconstruction and autonomous driving and compare them with WayveScenes101.

\subsection{Scene Reconstruction Datasets}
\label{subsec:scene-rec-datasets}

Many existing datasets for novel view synthesis consist of object-centric, outside-in images \cite{mildenhall2021nerf, wu2023omniobject3d, Knapitsch2017, barron2022mipnerf360, barron2021mip}, providing rich information for 3d reconstruction. More complex 3D indoor scenes \cite{dai2017scannet, yeshwanth2023scannet++, Knapitsch2017} are also widely used for novel view synthesis tasks. However, these datasets are mostly static and don't reflect the aforementioned challenges in reconstructing large-scale outdoor scenes. BlendedMVS \cite{yao2020blendedmvs} is another popular synthetic dataset consisting of diverse indoor and outdoor scenes with a wide range of textures and complexities was originally designed for multi-view stereo (MVS) reconstruction and is now also used for novel view synthesis. Although it captures large outdoor scenes, the lack of dynamic objects is a significant limitation.

\subsection{Driving Datasets}
\label{subsec:av-datasets}

In this section, we discuss existing driving datasets, focusing specifically on outdoor datasets suitable for novel view synthesis. It is important to note that these datasets were generally designed for multiple tasks, and consequently, they provide a rich variety of sensor modalities but often lack stereo views or a dedicated camera for testing novel view generation.

The Waymo Open Dataset \cite{Sun_2020_CVPR} features a wide field of view with five cameras. However, there is only a small baseline offset between cameras, and the spatial overlap of camera frustums is very small. Similarly, the NuScenes \cite{nuscenes2019} dataset provides a 360-degree view around the recording vehicle, but there is limited overlap between the cameras. KITTI-360 \cite{Liao2021ARXIV} offers a 360-degree view using two perspective cameras and two fisheye cameras, representing a significant improvement over the previous version since KITTI \cite{geiger2012we} did not include fisheye cameras. KITTI-360 also suggests a benchmark for novel view synthesis. Frames used for training and held-out frames for evaluation are sampled from static scenes with a different drop rate. Argoverse 2 \cite{Argoverse2} provides a 360-degree view and includes two stereo cameras but it lacks a dedicated camera for testing novel view generation.

The definition of a ``segment" or ``scene" varies across datasets. NuScenes uses a fixed length of 20 seconds for each selected scene, similar to Waymo Open. The Argoverse 2 Sensor Dataset uses time-based segments with a segment length of 30 seconds. KITTI-360 splits data into scenes according to the accumulated driving distances: each scene contains approx. $\SI{200}{m}$ of driving. 

A detailed comparison between our WayveScenes101 dataset and the aforementioned datasets is provided in Tab.~\ref{tab:datasets}. 

\sisetup{range-phrase=-}  
\sisetup{range-units=single}

\begin{table*}[ht]
    \centering
    \normalsize
    \resizebox{\textwidth}{!}{
    \begin{tabular}{lccccccccp{5cm}}
        \toprule
        Dataset & Year & Scenes & Scene Definition & Images & RGB Cameras & COLMAP Poses & Off-axis Test Camera & Scene Metadata & Locations \\
        \midrule
        KITTI~\cite{geiger2012we} & 2012 & 22 & Varying & 15K & 2 & \ding{55} & \ding{55} & \ding{55} & Karlsruhe \\
        KITTI-360~\cite{Liao2021ARXIV} & 2021 & 11 & Varying & 300K & 4 & \ding{55} & \ding{55}* & \ding{55} & Karlsruhe \\
        Waymo Open~\cite{Sun_2020_CVPR} & 2019 & 1150 & $\SI{20}{s}$ & 1M & 5 & \ding{55} & \ding{55} & \ding{55} & USA \\
        nuScenes~\cite{nuscenes2019} & 2019 & 1000 & $\SI{20}{s}$ & 1.4M & 6 & \ding{55} &  \ding{55} & \ding{55} & Boston, Singapore \\
        Argoverse~\cite{chang2019argoverse} & 2019 & 113 & $\SIrange{15}{30}{m}$ & 490K & 9 & \ding{55} & \ding{55}* & \ding{55} & USA \\
        Argoverse 2~\cite{Argoverse2} & 2023 & 1000 & $\SI{30}{s}$ & 2.7M & 9 & \ding{55} & \ding{55}* & \ding{55} & Austin, Detroit, Miami, Palo Alto, Pittsburgh, and Washington D.C \\
        \midrule
        WayveScenes101 (ours) & 2024 & 101 & $\SI{20}{s}$ & 101K & 5 & \checkmark & \checkmark & \checkmark & London, San Francisco Bay Area \\
        \bottomrule
    \end{tabular}}
    \caption{Comparison of popular AV datasets for evaluating novel view synthesis models. The asterisk (*) denotes datasets that do not have a dedicated off-axis camera but do have stereo cameras.}
    \label{tab:datasets}
\end{table*}

\section{Dataset}
\label{sec:dataset}

\begin{figure*}[ht]
    \centering
    \begin{minipage}{0.24\textwidth}
        \centering
        \includegraphics[width=\textwidth]{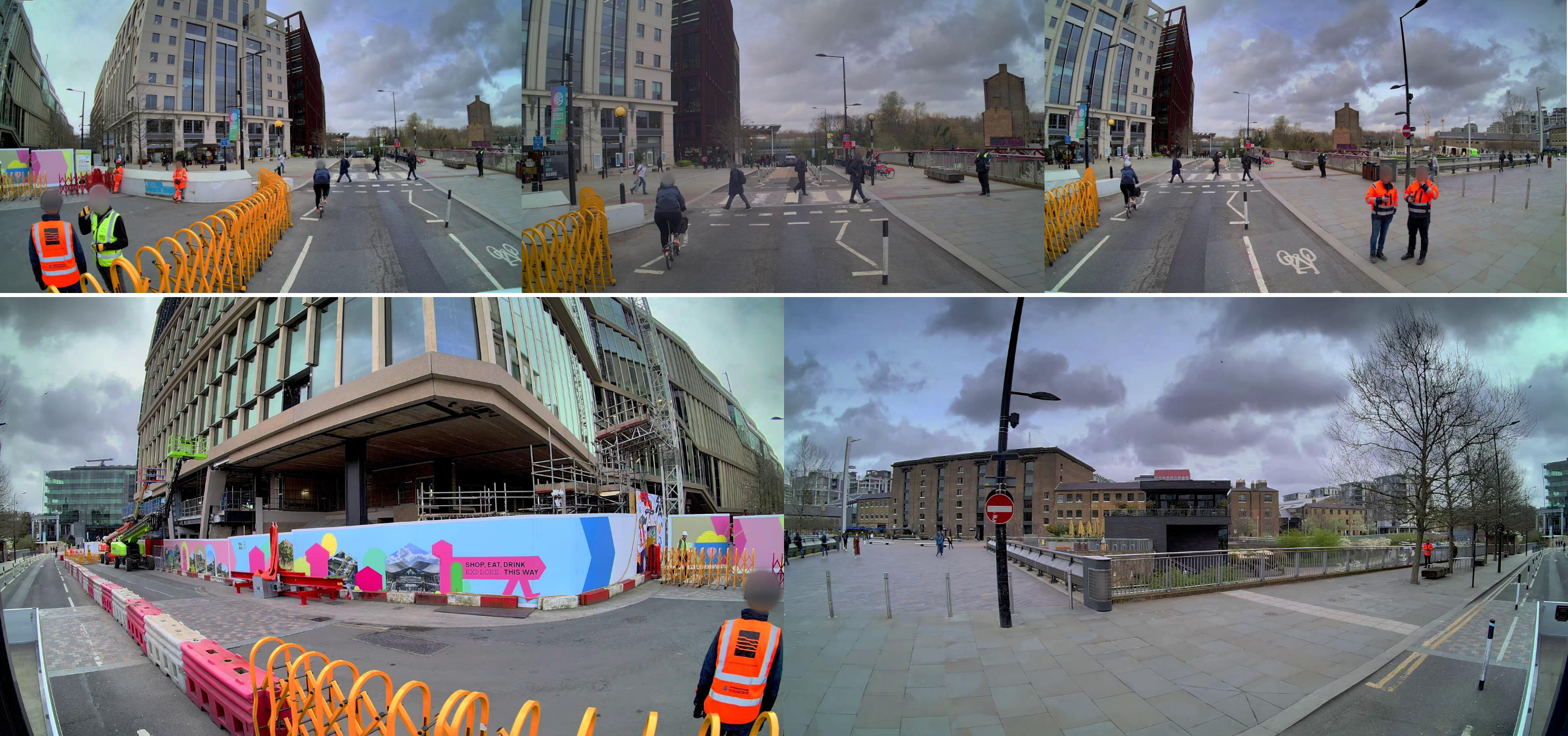}
        \label{fig:subfig0}
    \end{minipage}
    \hfill
    \begin{minipage}{0.24\textwidth}
        \centering
        \includegraphics[width=\textwidth]{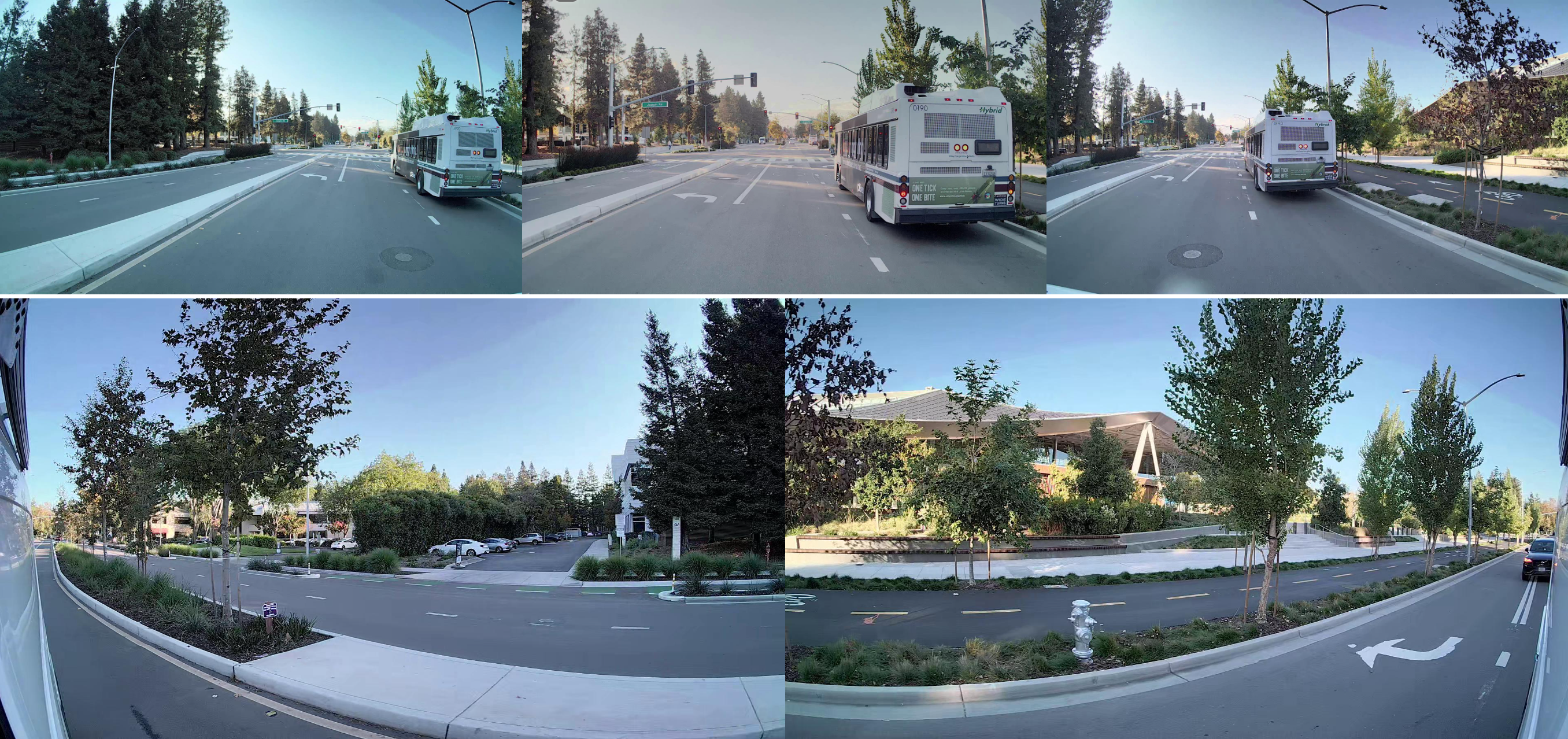}
        \label{fig:subfig1}
    \end{minipage}
    \hfill
    \begin{minipage}{0.24\textwidth}
        \centering
        \includegraphics[width=\textwidth]{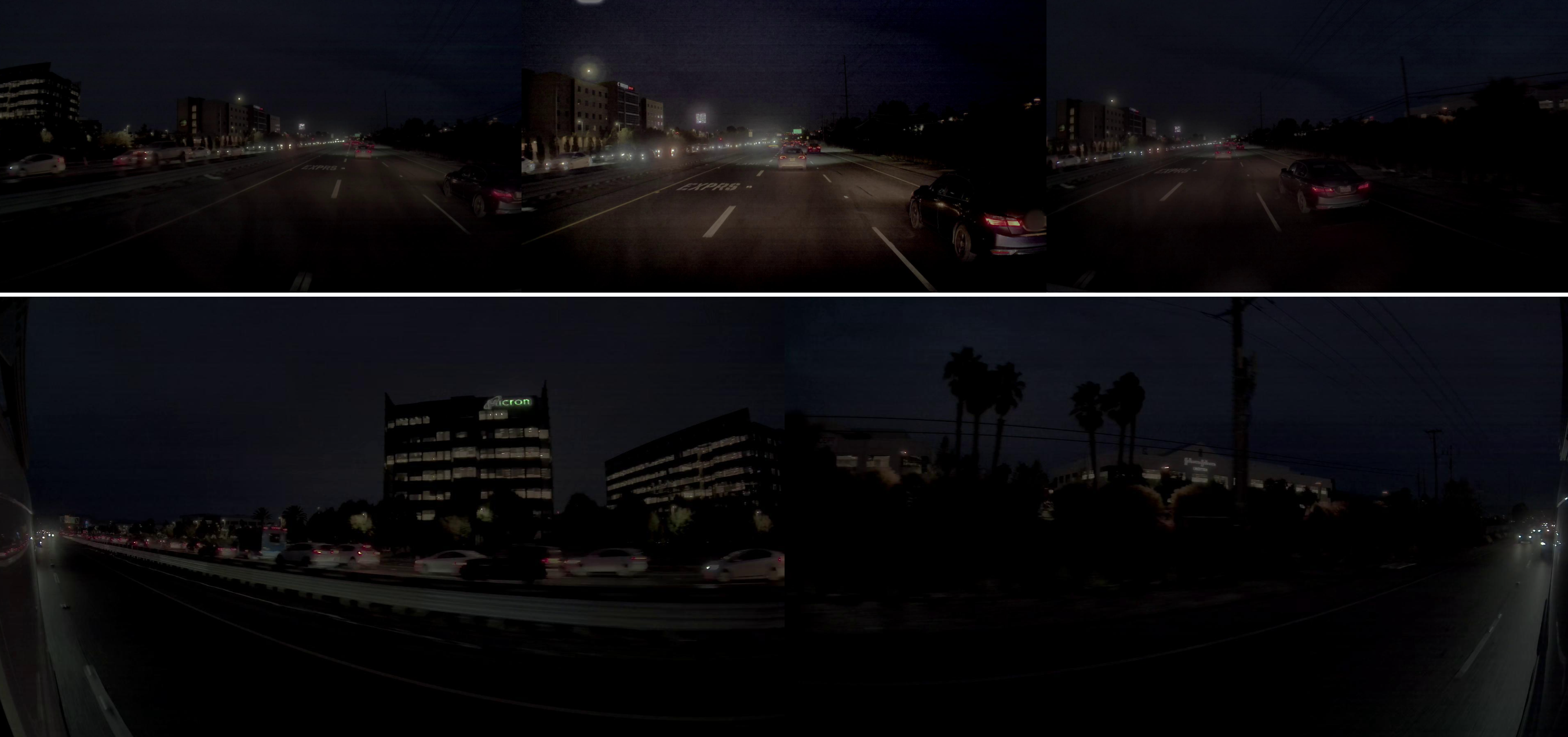}
        \label{fig:subfig2}
    \end{minipage}
    \hfill
    \begin{minipage}{0.24\textwidth}
        \centering
        \includegraphics[width=\textwidth]{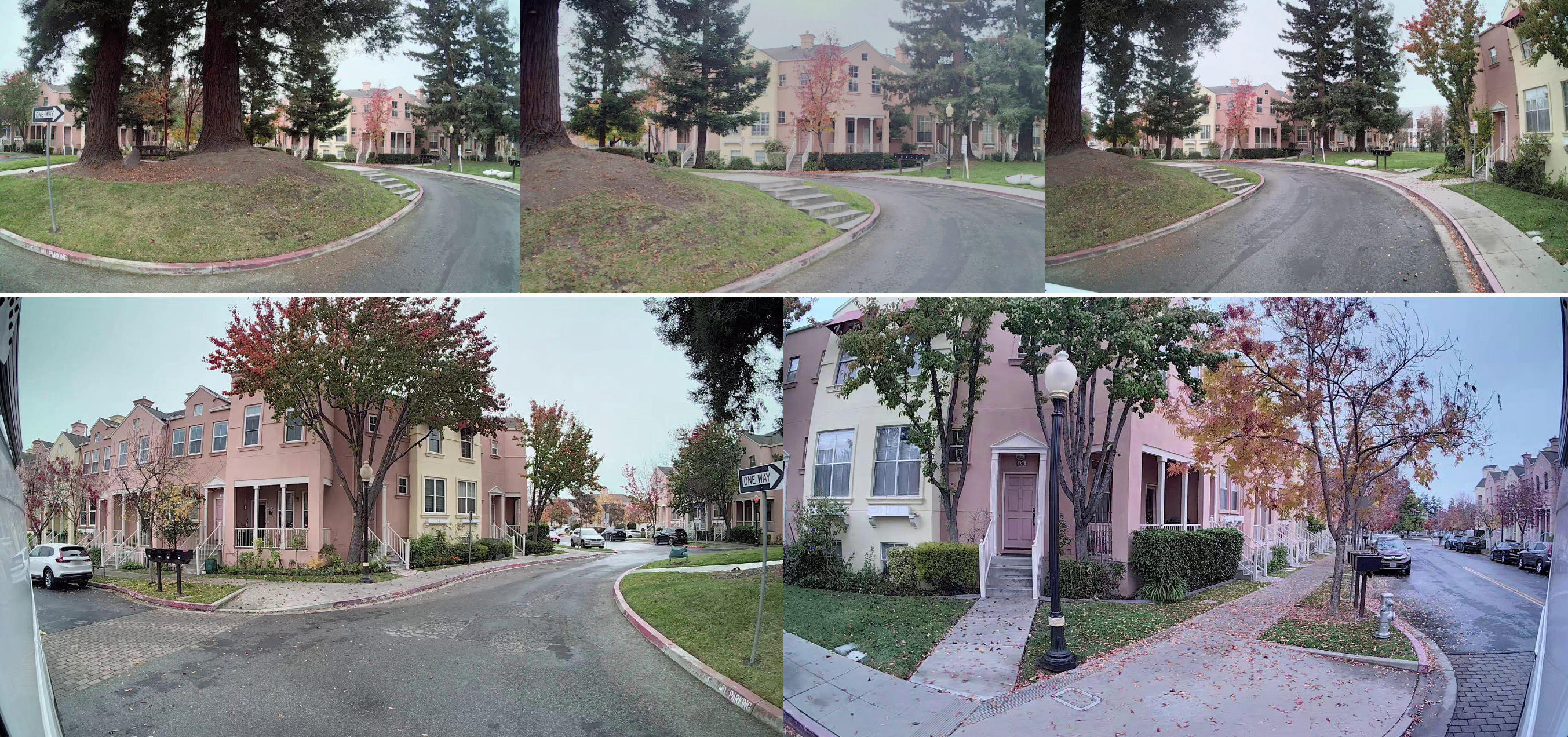}
        \label{fig:subfig3}
    \end{minipage}
    \begin{minipage}{0.24\textwidth}
        \centering
        \includegraphics[width=\textwidth]{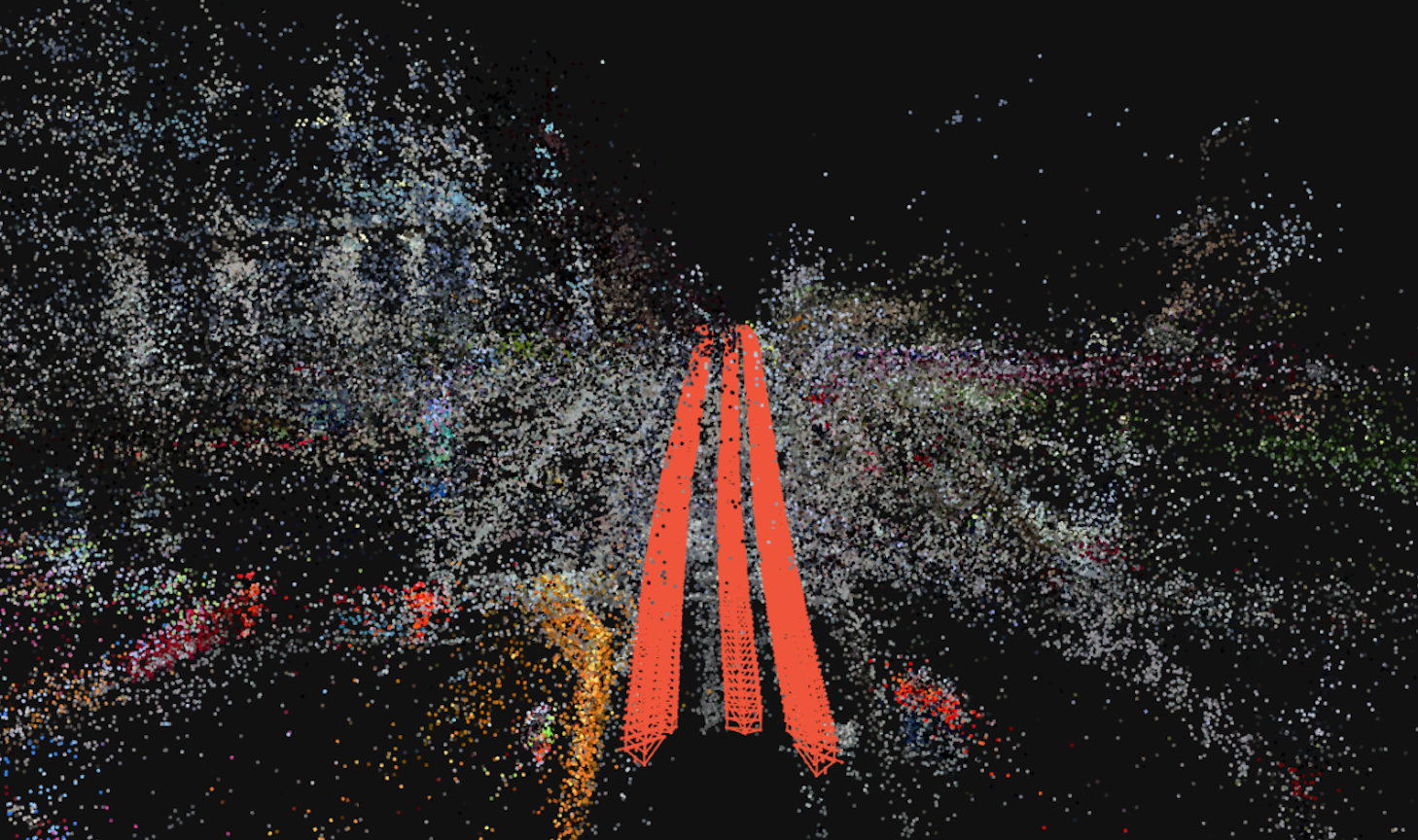}
        \subcaption{\texttt{scene\_024}}
        \label{fig:subfig4}
    \end{minipage}
    \hfill
    \begin{minipage}{0.24\textwidth}
        \centering
        \includegraphics[width=\textwidth]{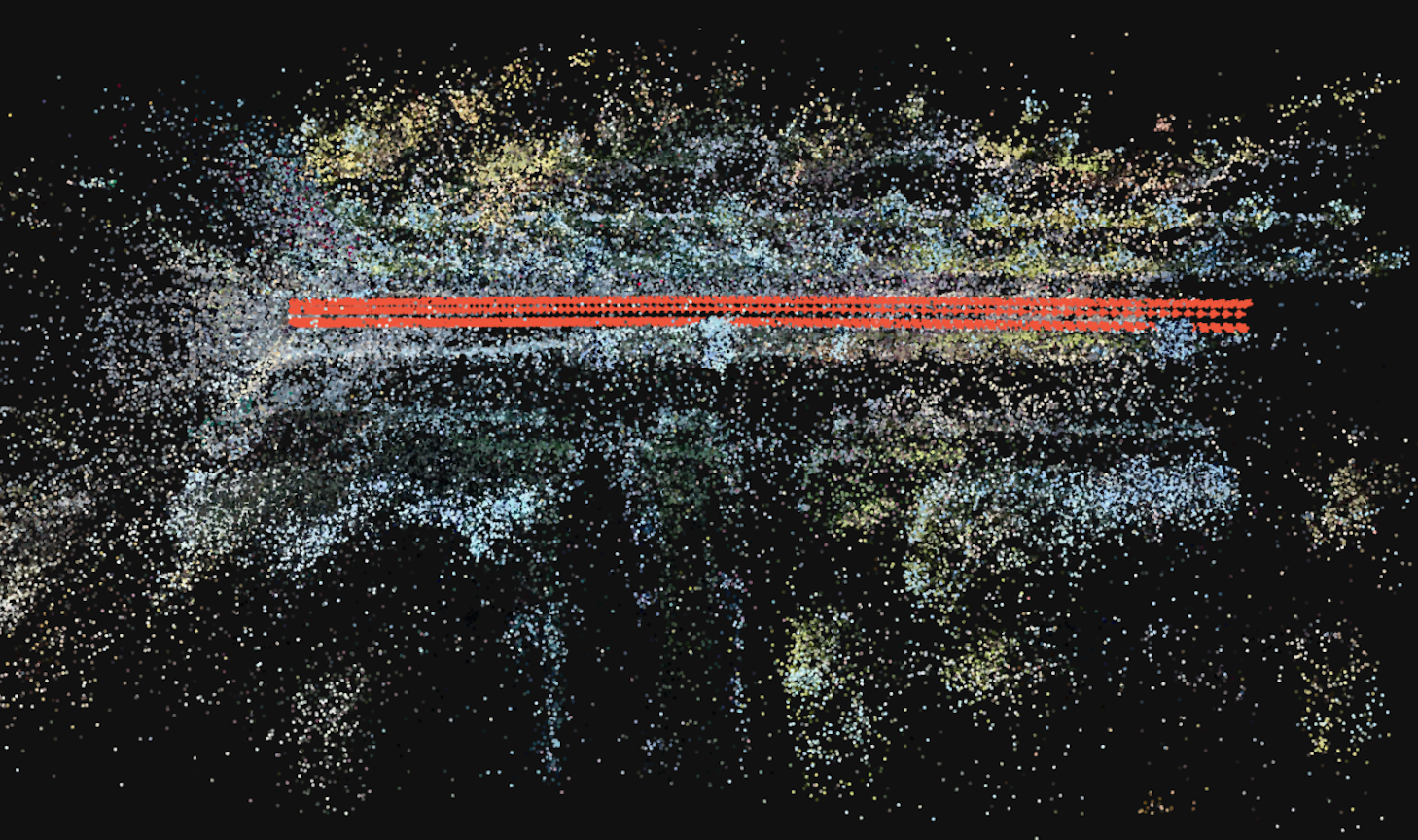}
        \subcaption{\texttt{scene\_032}}
        \label{fig:subfig5}
    \end{minipage}
    \hfill
    \begin{minipage}{0.24\textwidth}
        \centering
        \includegraphics[width=\textwidth]{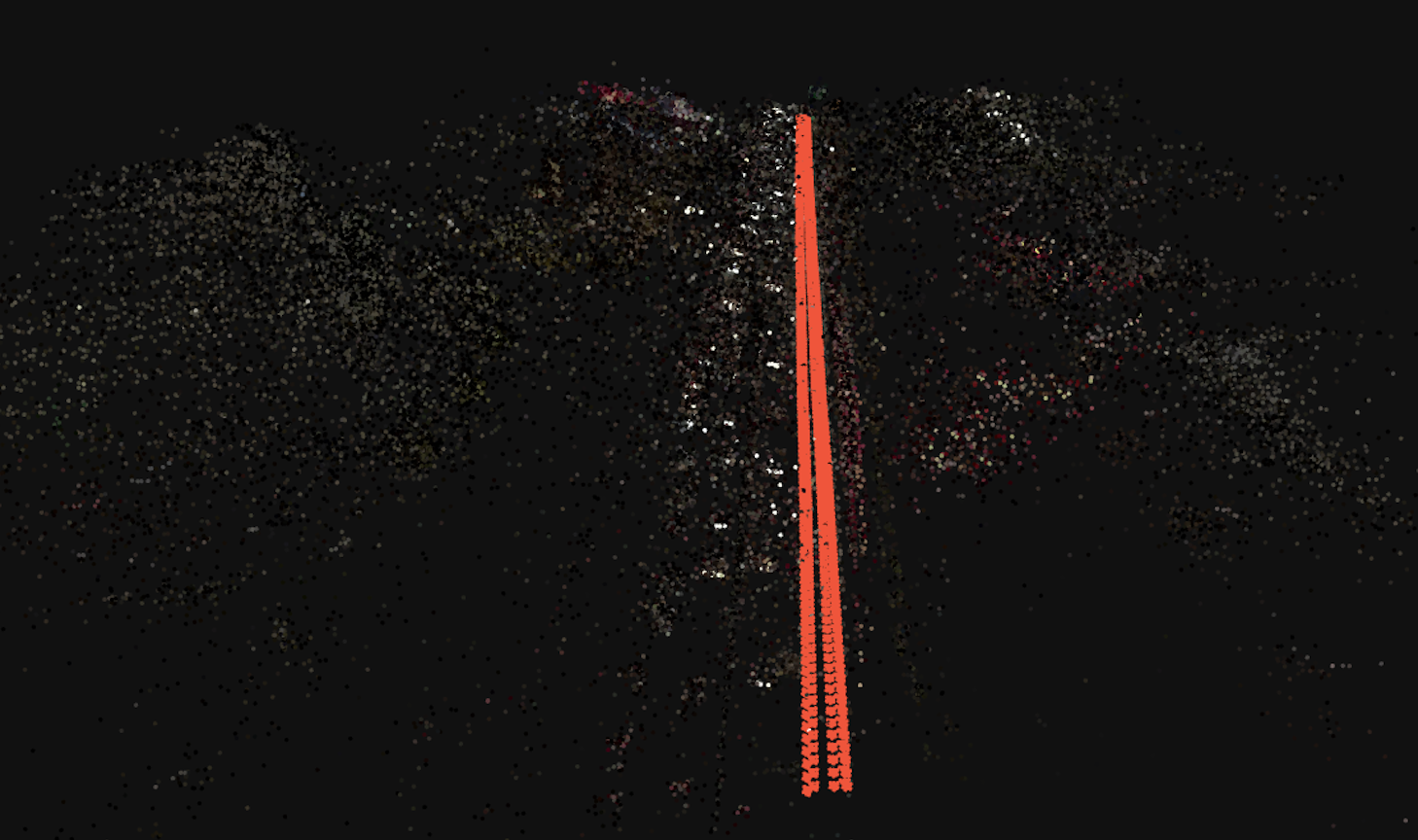}
        \subcaption{\texttt{scene\_071}}
        \label{fig:subfig6}
    \end{minipage}
    \hfill
    \begin{minipage}{0.24\textwidth}
        \centering
        \includegraphics[width=\textwidth]{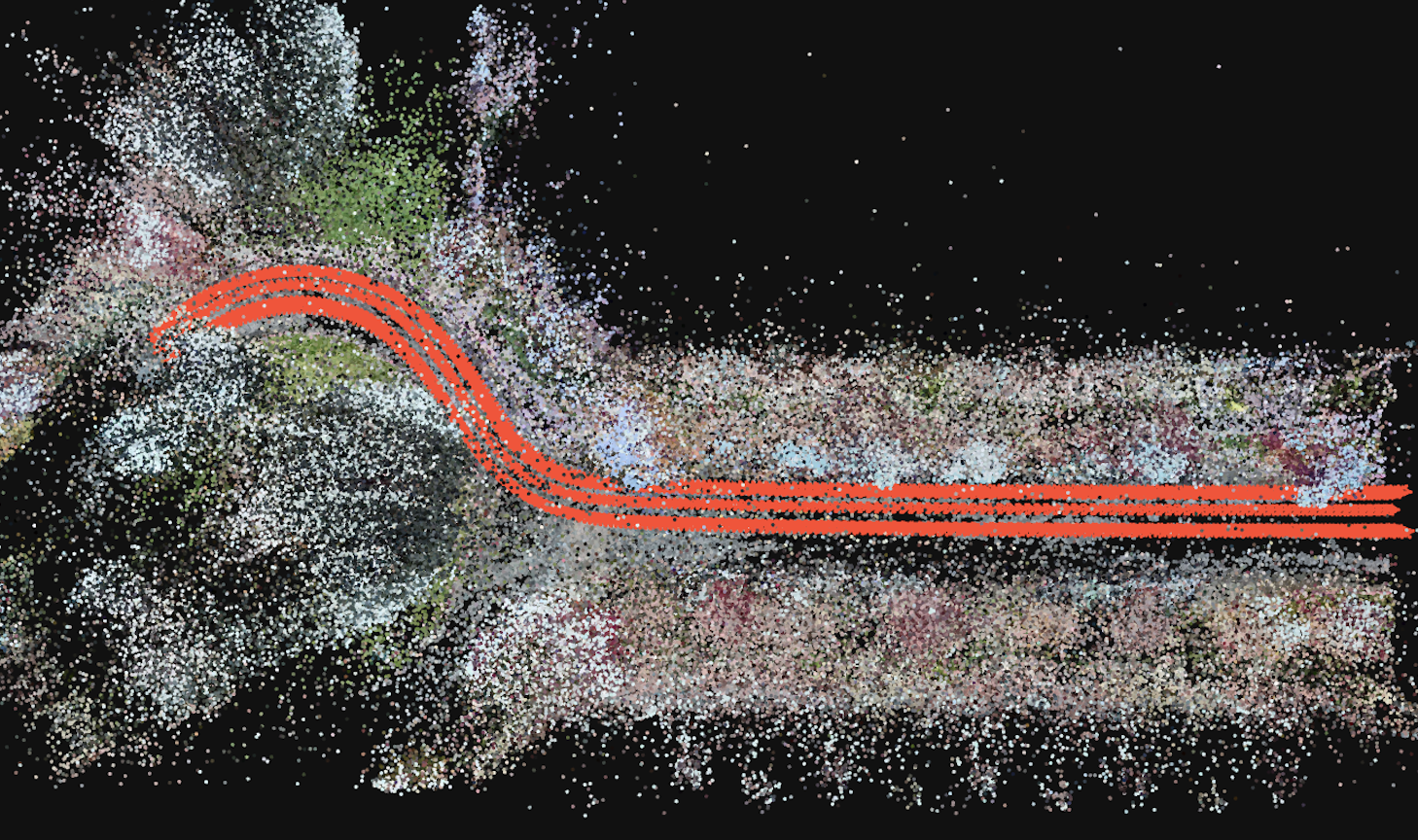}
        \subcaption{\texttt{scene\_100}}
        \label{fig:subfig7}
    \end{minipage}
    \vspace{5mm}

    \caption{We visualise exemplary scenes from our WayveScenes101 dataset. The top row shows all camera images for a single time step. The bottom row illustrates the 3D camera poses and 3D keypoints for the given scene.}
    \label{fig:scenes-overview}
\end{figure*}

Our publicly available WayveScenes101 dataset provides 101 scenes from a variety of driving environments, such as urban, suburban, and highways, across different weather and lighting conditions. Each 20 seconds scene includes time-synchronised views from five vehicle-mounted cameras and associated camera poses obtained from COLMAP~\cite{schoenberger2016sfm}. We also provide masks for each image that indicate image regions that should not be used for scene reconstruction (blurred faces, license plates, and ego-vehicle masks). We illustrate example scenes in Fig.~\ref{fig:scenes-overview}.

\subsection{Scene Selection}
\label{subsec:scene-selection}

Our selected scenes are chosen to be representative of the challenges faced when deploying novel view synthesis methods in the real world, such as dynamically changing traffic lights, sun glare, or moving pedestrians. For a detailed breakdown of the metadata distribution for all scenes, please refer to Fig.~\ref{fig:metadata}. Furthermore, we aim to maximize the diversity of ego-vehicle driving behaviour across scenes (as the ego-vehicle velocity and trajectory depend on the traffic situation and environment). As a result, the distance travelled in a scene varies from less than $\SI{20}{m}$ to more than $\SI{500}{m}$ (see Tab.~\ref{fig:distance_distribution}).

We also provide detailed scene metadata for customized evaluation of novel view synthesis models in specific science, i.e. scenes with rain, fully static scenes, or nighttime scenes. This metadata may be used to calculate specific model performance metrics for specific scene types.

\subsection{Camera Rig}
\label{subsec:camera-rig}

\begin{figure}
  \centering
  \includegraphics[width=\linewidth]{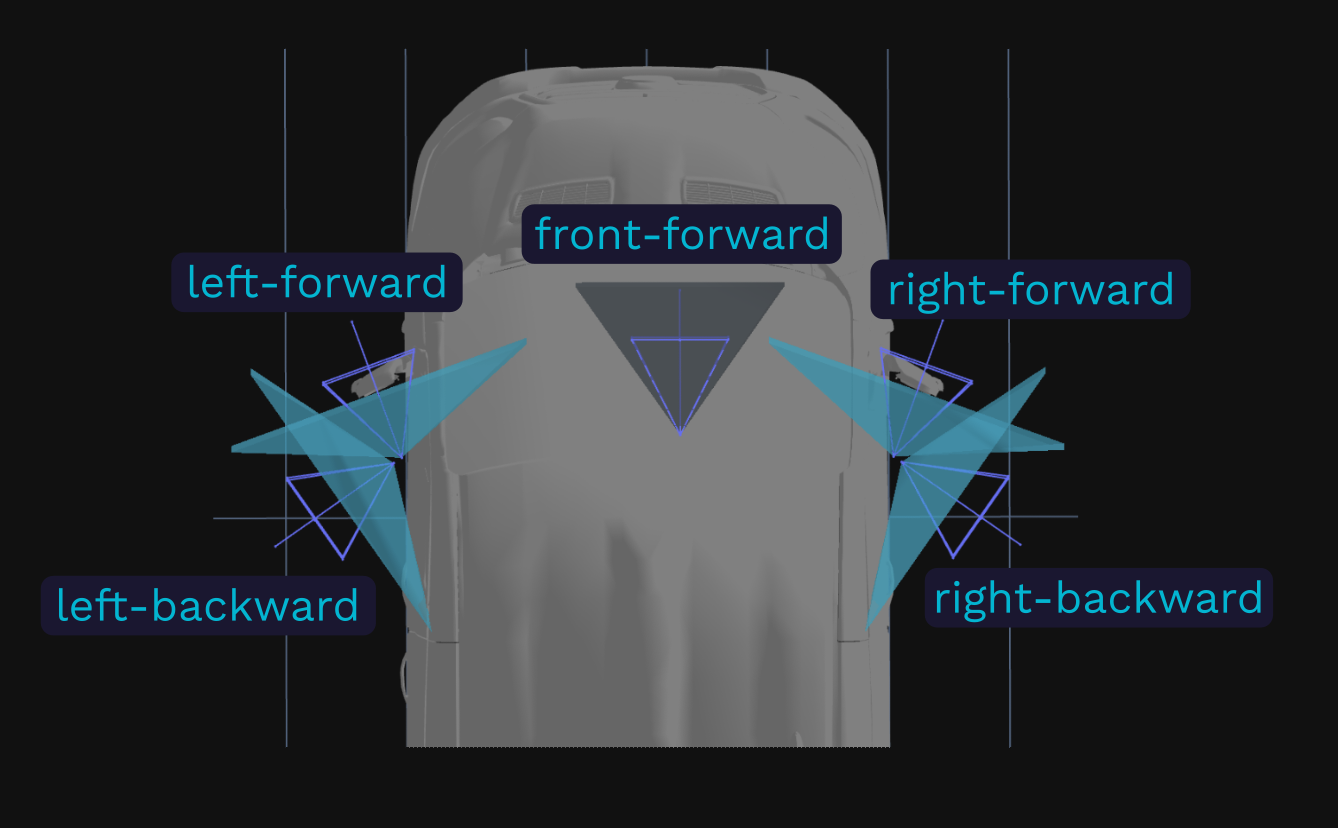}
  \caption{Camera rig overview.}
  \label{fig:camera-rig}
\end{figure}

We visualise the camera rig used to record the dataset in Fig.~\ref{fig:camera-rig}. Our rig consists of four wide angle cameras (\texttt{left-backward}, \texttt{left-forward}, \texttt{right-forward}, \texttt{right-backward}) and one centered forward-facing camera with a smaller field-of-view (\texttt{front-forward}). We list the camera parameters in Tab.~\ref{tab:camera_specs}.

We provide an evaluation protocol for a held-out evaluation camera to specifically measure off-axis reconstruction quality, which is crucial for estimating the generalisation capabilities of novel view synthesis models. We propose to use the \texttt{front-forward} camera for model evaluation as it has a complete frustum overlap with the lateral forward-facing cameras but maintains a significant baseline distance from them ($\SI{\sim0.9}{m}$ and $\SI{\sim1.2}{m}$, respectively). The large baseline requires scene reconstruction models to model the geometry of the scene with high accuracy in order to account for the shift in perspective when evaluating on the evaluation camera. A frame rate of 10 frames per second for each camera allows for accurate scene reconstruction, in particular for dynamic scenes.

\begin{table}[!tbp]
\begin{center}
\begin{tabular}{rcc}
\toprule
& \texttt{ff} & \texttt{lf},\texttt{rf},\texttt{lb},\texttt{rb} \\
\midrule
Resolution & 1920×1080 & 1920×1080 \\
% Horizontal FOV & $\pm31\degree$ & $\pm45\degree$ \\
% Vertical FOV & $\pm18.5\degree$ & $\pm29\degree$ \\
Nominal H-FOV & $70\degree$ & $120\degree$ \\
\bottomrule
\end{tabular}
\end{center}
\caption{\small{Details for \texttt{front-forward} (\texttt{ff}), \texttt{left-forward} (\texttt{lf}), \texttt{right-forward} (\texttt{rf}), \texttt{left-backward} (\texttt{lb}), and \texttt{right-backward} (\texttt{rb}) cameras.}}
\label{tab:camera_specs}
\end{table}

\section{Tasks}
\label{sec:tasks}

Our WayveScenes101 dataset is generally designed for evaluating novel scene reconstruction methods on driving scenes. Within this task, we place a particular focus on measuring the off-axis reconstruction quality of these methods. To enable this, we propose a specific evaluation protocol that leverages the \texttt{front-forward} camera which should not used for model training. As per Sec.~\ref{subsec:camera-rig}, the \texttt{front-forward} camera sits between the \texttt{left-forward} and \texttt{right-forward} cameras and has a large frustum overlap with these cameras, allowing it to be used to measure off-axis scene reconstruction model performance (see Subsec.~\ref{subsec:camera-rig} for details).

\subsection{Model Evaluation Protocol}
\label{subsec:model_evaluation}

We use peak-signal-noise ratio (PSNR), structural similarity index measure (SSIM), learned perceptual image patch similarity (LPIPS) and the Fréchet inception distance (FID) to quantify model performance. We calculate the PSNR, SSIM, and LPIPS metrics for all images separately. A given metric $M$ is averaged over all images in a given scene $S$ as per Equation \ref{eq:metric_agg_scene}:

\begin{equation} \label{eq:metric_agg_scene}
    M_{S} = \frac{1}{|S|} \sum_{\text{Image} \in S} M_{\text{Image}}
\end{equation}

The FID metric is calculated over all predicted and ground-truth images in one scene in one sweep as it compares image distributions rather than individual images~\cite{heusel2017gans}. Considering our proposed off-axis evaluation camera (\texttt{front-forward}), the metrics can be split into training metrics by not including the \texttt{front-forward} camera in $S$, and testing metrics by only including the \texttt{front-forward} camera in $S$.

To obtain the metrics for the full dataset $\mathcal{D}$, we average the metrics over all scenes $S$ where every scene is weighted with the same weight:

\begin{equation}
    M_{\mathcal{D}} = \frac{1}{|\mathcal{D}|} \sum_{S \in \mathcal{D}} M_{S}.
\end{equation}
\label{eq:metric_agg}

We also propose to evaluate models for specific subsets of scenes, i.e. all nighttime scenes or only fully static scenes. In this case, we sample $S$ from a given subset $\{S_i\} \subset \mathcal{D}$ and average over the number of scenes in $\{S_i\}$.

% We propose a meta-metric to combine previously mentioned metrics into a single scalar value to allow for easy model ranking. We define this metric as the average of the four previously described metrics. While the SSIM, LPIPS, and FID metrics are bound in the unit interval $[0,1]$, PSNR values lie in $[0, \infty)$. To map the PSNR metric to the unit interval, we propose a Sigmoid mapping of PSNR values, where we define a PSNR value of 10 to map to 0.1 (bad reconstruction) and a PSNR value of 30 (good reconstruction) to 0.9, resulting in the mapping function

% \begin{equation}
%     \text{PSNR}_{\text{unit}} = \frac{1}{1 + \exp(-k * (\text{PSNR} - 20))},
% \end{equation}

% where $k=\frac{\log(9)}{10}$. 

% Finally, we obtain the meta-metric $M$ by averaging over the four proposed metrics:

% \begin{equation}
%     M = \frac{1}{4} (\text{PSNR}_\text{unit} + \text{SSIM} + (1 -\text{LPIPS}) + (1 - \text{FID})),
% \end{equation}

% where $M \rightarrow 1$ defines perfect reconstruction quality and $M \rightarrow 0$ defines the worst reconstruction quality.

\section{Conclusion}
\label{sec:conclusion}

We presented our novel WayveScenes101 dataset designed to measure the performance of novel scene synthesis methods for driving scenes. We showed that our dataset provides a wide range of scenes with various environmental conditions and driving scenarios. We described the camera rig that was used to record the scenes and detailed how we envision reconstruction models to be evaluated on our dataset. 

{\small
\bibliographystyle{ieee_fullname}
\bibliography{main}
}

\clearpage

\onecolumn

\begin{center}
\Large{\bf Appendix}\\
\vspace{0.4cm}
\end{center}

\setcounter{section}{0}
\setcounter{equation}{0}
\setcounter{figure}{0}
\setcounter{table}{0}
\makeatletter

\renewcommand{\thesection}{S.\arabic{section}}
\renewcommand{\thesubsection}{S.\arabic{section}.\arabic{subsection}}
\renewcommand{\thetable}{S.\arabic{table}}
\renewcommand{\thefigure}{S.\arabic{figure}}

\normalsize

\definecolor{wayve_blue}{RGB}{3,181,209}
% \definecolor{green}{RGB}{0,204,102}
% \definecolor{red}{RGB}{255,51,51}

\begin{figure}[ht]
    \centering
    % First Row
    \begin{subfigure}[b]{0.45\textwidth}
        \centering
        \begin{tikzpicture}
            \begin{axis}[
                title=\textbf{Weather},
                xbar,
                ytick=data,
                symbolic y coords={Rainy, Overcast, Sunny},
                axis y line*=left,
                axis x line*=bottom,
                width=\textwidth,
                height=3cm,
                enlarge y limits  = 0.2,
                enlarge x limits  = 0.02,
                xmin=0, % Start the x-axis at 0
                xmax=100, % Adjust the x-axis limit to fit the data
                ytick style={draw=none}, % Remove y-tick marks
                axis line style={draw=none},
                xlabel={Count},
                % bar width=10pt, % Adjust bar width
                yticklabel style={font=\small},
                xticklabel style={font=\small},
                % yticklabel style={font=\normalsize}, % Adjust y-label style
                y dir=reverse,
            ]
            \addplot[fill=wayve_blue] coordinates { (63,Sunny) (29,Overcast)(9,Rainy)};
            \end{axis}
        \end{tikzpicture}
    \end{subfigure}
    \hfill
    \begin{subfigure}[b]{0.45\textwidth}
        \centering
        \begin{tikzpicture}
            \begin{axis}[
                title=\textbf{Time of Day},
                xbar,
                ytick=data,
                symbolic y coords={Day, Night, Dusk},
                axis y line*=left,
                axis x line*=bottom,
                width=\textwidth,
                height=3cm,
                enlarge y limits  = 0.2,
                enlarge x limits  = 0.02,
                xmin=0, % Start the x-axis at 0
                xmax=100, % Adjust the x-axis limit to fit the data
                axis line style={draw=none},
                tick style={draw=none},
                xlabel={Count},
                yticklabel style={font=\small},
                xticklabel style={font=\small},
            ]
            \addplot[fill=wayve_blue] coordinates {(94,Day) (5,Night) (2,Dusk) };
            \end{axis}
        \end{tikzpicture}
    \end{subfigure}
    
    % Second Row
    % \todo{I made the ``Same Direction Vehicle Traffic'' Yes/No plot more compact which I think looks better. Wdyt? Paul: I like.}
    \vspace{1em}
    \begin{subfigure}[b]{0.45\textwidth}
        \centering
        \begin{tikzpicture}
            \begin{axis}[
                title=\textbf{Same Direction Vehicle Traffic},
                xbar,
                ytick=data,
                symbolic y coords={Yes, No},
                axis y line*=left,
                axis x line*=bottom,
                width=\textwidth,
                height=2.8cm,
                enlarge y limits  = 0.8,
                enlarge x limits  = 0.02,
                xmin=0, % Start the x-axis at 0
                xmax=100, % Adjust the x-axis limit to fit the data
                axis line style={draw=none},
                tick style={draw=none},
                xlabel={Count},
                yticklabel style={font=\small},
                xticklabel style={font=\small},
            ]
            \addplot[fill=wayve_blue] coordinates {(66,Yes) (35,No)};
            % \addplot[fill=green] coordinates {(66,Yes)};
            % \addplot[fill=red] coordinates {(35,No)};
            \end{axis}
        \end{tikzpicture}
    \end{subfigure}
    \hfill
    \begin{subfigure}[b]{0.45\textwidth}
        \centering
        \begin{tikzpicture}
            \begin{axis}[
                title=\textbf{Road Type},
                xbar,
                ytick=data,
                symbolic y coords={Urban, Residential, Rural, Highway},
                axis y line*=left,
                axis x line*=bottom,
                width=\textwidth,
                height=3.8cm,
                enlarge y limits  = 0.2,
                enlarge x limits  = 0.1,
                xmin=0, % Start the x-axis at 0
                xmax=100, % Adjust the x-axis limit to fit the data
                axis line style={draw=none},
                tick style={draw=none},
                xlabel={Count},
                yticklabel style={font=\small, anchor=east, xshift=15pt},
                xticklabel style={font=\small},
            ]
            \addplot[fill=wayve_blue] coordinates {(51,Urban) (35,Residential) (9,Rural) (6,Highway)};
            \end{axis}
        \end{tikzpicture}
    \end{subfigure}

    % Third Row
    \vspace{1em}
    \begin{subfigure}[b]{0.45\textwidth}
        \centering
        \begin{tikzpicture}
            \begin{axis}[
                title=\textbf{Oncoming Vehicle Traffic},
                xbar,
                ytick=data,
                symbolic y coords={Yes, No},
                axis y line*=left,
                axis x line*=bottom,
                width=\textwidth,
                height=2.8cm,
                enlarge y limits  = 0.8,
                enlarge x limits  = 0.02,
                xmin=0, % Start the x-axis at 0
                xmax=100, % Adjust the x-axis limit to fit the data
                axis line style={draw=none},
                tick style={draw=none},
                xlabel={Count},
                yticklabel style={font=\small},
                xticklabel style={font=\small},
            ]
            \addplot[fill=wayve_blue] coordinates {(61,Yes) (40,No)};
            \end{axis}
        \end{tikzpicture}
    \end{subfigure}
    \hfill
    \begin{subfigure}[b]{0.45\textwidth}
        \centering
        \begin{tikzpicture}
            \begin{axis}[
                title=\textbf{Crossing Vehicle Traffic},
                xbar,
                ytick=data,
                symbolic y coords={Yes, No},
                axis y line*=left,
                axis x line*=bottom,
                width=\textwidth,
                height=2.8cm,
                enlarge y limits  = 0.8,
                enlarge x limits  = 0.02,
                xmin=0, % Start the x-axis at 0
                xmax=100, % Adjust the x-axis limit to fit the data
                axis line style={draw=none},
                tick style={draw=none},
                xlabel={Count},
                yticklabel style={font=\small},
                xticklabel style={font=\small},
            ]
            \addplot[fill=wayve_blue] coordinates {(75,Yes) (26,No)};
            \end{axis}
        \end{tikzpicture}
    \end{subfigure}

    % Fourth Row
    \vspace{1em}
    \begin{subfigure}[b]{0.45\textwidth}
        \centering
        \begin{tikzpicture}
            \begin{axis}[
                title=\textbf{Pedestrian Crossing Road},
                xbar,
                ytick=data,
                symbolic y coords={Yes, No},
                axis y line*=left,
                axis x line*=bottom,
                width=\textwidth,
                height=2.8cm,
                enlarge y limits  = 0.8,
                enlarge x limits  = 0.02,
                xmin=0, % Start the x-axis at 0
                xmax=100, % Adjust the x-axis limit to fit the data
                axis line style={draw=none},
                tick style={draw=none},
                xlabel={Count},
                yticklabel style={font=\small},
                xticklabel style={font=\small},
            ]
            \addplot[fill=wayve_blue] coordinates {(80,Yes) (21,No)};
            \end{axis}
        \end{tikzpicture}
    \end{subfigure}
    \hfill
    \begin{subfigure}[b]{0.45\textwidth}
        \centering
        \begin{tikzpicture}
            \begin{axis}[
                title=\textbf{Pedestrians on Sidewalk},
                xbar,
                ytick=data,
                symbolic y coords={Yes, No},
                axis y line*=left,
                axis x line*=bottom,
                width=\textwidth,
                height=2.8cm,
                enlarge y limits  = 0.8,
                enlarge x limits  = 0.02,
                xmin=0, % Start the x-axis at 0
                xmax=100, % Adjust the x-axis limit to fit the data
                axis line style={draw=none},
                tick style={draw=none},
                xlabel={Count},
                yticklabel style={font=\small},
                xticklabel style={font=\small},
            ]
            \addplot[fill=wayve_blue] coordinates {(45,Yes) (56,No)};
            \end{axis}
        \end{tikzpicture}
    \end{subfigure}

    % Fifth Row
    \vspace{1em}
    \begin{subfigure}[b]{0.45\textwidth}
        \centering
        \begin{tikzpicture}
            \begin{axis}[
                title=\textbf{Cyclists / Motorbikes Present},
                xbar,
                ytick=data,
                symbolic y coords={Yes, No},
                axis y line*=left,
                axis x line*=bottom,
                width=\textwidth,
                height=2.8cm,
                enlarge y limits  = 0.8,
                enlarge x limits  = 0.02,
                xmin=0, % Start the x-axis at 0
                xmax=100, % Adjust the x-axis limit to fit the data
                axis line style={draw=none},
                tick style={draw=none},
                xlabel={Count},
                yticklabel style={font=\small},
                xticklabel style={font=\small},
            ]
            \addplot[fill=wayve_blue] coordinates {(15,Yes) (86,No)};
            \end{axis}
        \end{tikzpicture}
    \end{subfigure}
    \hfill
    \begin{subfigure}[b]{0.45\textwidth}
        \centering
        \begin{tikzpicture}
            \begin{axis}[
                title=\textbf{Large Exposure Change},
                xbar,
                ytick=data,
                symbolic y coords={Yes, No},
                axis y line*=left,
                axis x line*=bottom,
                width=\textwidth,
                height=2.8cm,
                enlarge y limits  = 0.8,
                enlarge x limits  = 0.02,
                xmin=0, % Start the x-axis at 0
                xmax=100, % Adjust the x-axis limit to fit the data
                axis line style={draw=none},
                tick style={draw=none},
                xlabel={Count},
                yticklabel style={font=\small},
                xticklabel style={font=\small},
            ]
            \addplot[fill=wayve_blue] coordinates {(63,Yes) (38,No)};
            \end{axis}
        \end{tikzpicture}
    \end{subfigure}
    \end{figure}

    % Sixth Row
    \vspace{1em}
    \begin{figure}
    \begin{subfigure}[b]{0.45\textwidth}\ContinuedFloat
        \centering
        \begin{tikzpicture}
            \begin{axis}[
                title=\textbf{Traffic Light Change},
                xbar,
                ytick=data,
                symbolic y coords={Yes, No},
                axis y line*=left,
                axis x line*=bottom,
                width=\textwidth,
                height=2.8cm,
                enlarge y limits  = 0.8,
                enlarge x limits  = 0.02,
                xmin=0, % Start the x-axis at 0
                xmax=100, % Adjust the x-axis limit to fit the data
                axis line style={draw=none},
                tick style={draw=none},
                xlabel={Count},
                yticklabel style={font=\small},
                xticklabel style={font=\small},
            ]
            \addplot[fill=wayve_blue] coordinates {(9,Yes) (92,No)};
            \end{axis}
        \end{tikzpicture}
    \end{subfigure}
    \hfill
    \begin{subfigure}[b]{0.45\textwidth}
        \centering
        \begin{tikzpicture}
            \begin{axis}[
                title=\textbf{Changing Brake Light or Indicator},
                xbar,
                ytick=data,
                symbolic y coords={Yes, No},
                axis y line*=left,
                axis x line*=bottom,
                width=\textwidth,
                height=2.8cm,
                enlarge y limits  = 0.8,
                enlarge x limits  = 0.02,
                xmin=0, % Start the x-axis at 0
                xmax=100, % Adjust the x-axis limit to fit the data
                axis line style={draw=none},
                tick style={draw=none},
                xlabel={Count},
                yticklabel style={font=\small},
                xticklabel style={font=\small},
            ]
            \addplot[fill=wayve_blue] coordinates {(28,Yes) (73,No)};
            \end{axis}
        \end{tikzpicture}
    \end{subfigure}
    \caption{Scene metadata distribution}
    \label{fig:metadata}

\end{figure}
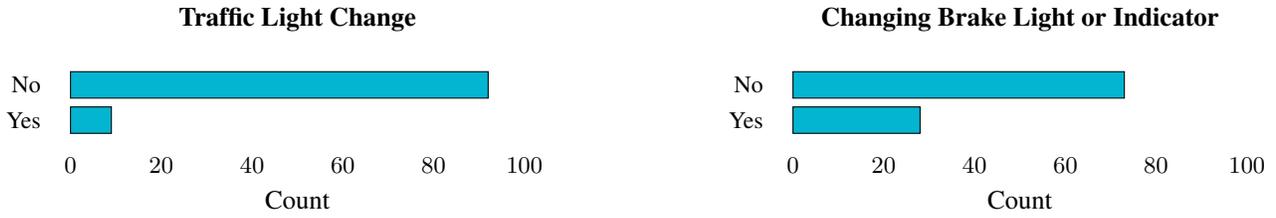

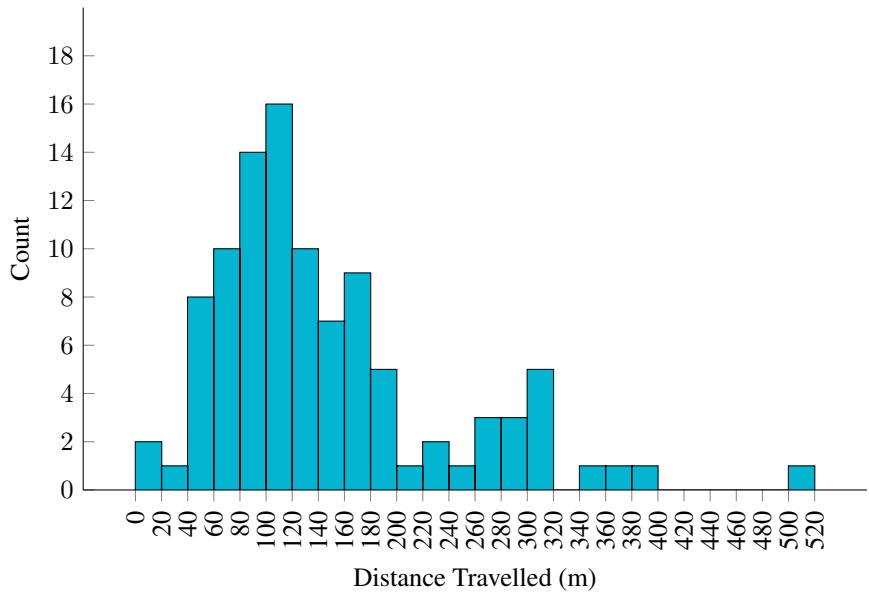
\begin{figure*}[ht]
    \centering
    \begin{tikzpicture}
        \begin{axis}[
            width=12cm,
            height=8cm,
            ybar,
            bar width=10pt,
            xtick={10, 30, 50, 70, 90, 110, 130, 150, 170, 190, 210, 230, 250, 270, 290, 310, 330, 350, 370, 390, 410, 430, 450, 470, 490, 510, 530},
            xticklabels={0, 20, 40, 60, 80, 100, 120, 140, 160, 180, 200, 220, 240, 260, 280, 300, 320, 340, 360, 380, 400, 420, 440, 460, 480, 500, 520},
            xticklabel style={rotate=90, anchor=east},
            xlabel={Distance Travelled (m)},
            ylabel={Count},
            ymin=0,
            ymax=20,
            ytick={0, 2, 4, 6, 8, 10, 12, 14, 16, 18},
            % nodes near coords,
            % every node near coord/.append style={font=\footnotesize, rotate=90, anchor=west}
            axis x line*=bottom, % Only show the bottom x-axis line
            axis y line*=left,   % Only show the left y-axis line
        ]
        \addplot[fill=wayve_blue] coordinates {(20,2) (40,1) (60,8) (80,10) (100,14) (120,16) (140,10) (160,7) (180,9) (200, 5) (220,1) (240,2) (260,1) (280,3) (300,3) (320,5) 
                                (340,0) (360,1) (380,1) (400,1) (420,0) (440,0) (460,0) (480,0) (500,0) (520,1)};
        \end{axis}
    \end{tikzpicture}
    \caption{Distribution of travelled distance per scene}
    \label{fig:distance_distribution}
\end{figure*}

\end{document}